\documentclass{esannV2}
\usepackage{graphicx}
\usepackage[latin1]{inputenc}
\usepackage{amssymb,amsmath,array}

\usepackage[usenames,dvipsnames]{xcolor}
\usepackage[colorlinks=true,urlcolor=teal,linkcolor=teal,citecolor=teal]{hyperref}
\usepackage{subcaption}
\DeclareMathOperator{\argmax}{arg\,max}

\begin{document}

\title{Is Q-learning an Ill-posed Problem?}

\author{Philipp Wissmann$^{1,2}$, Daniel Hein$^1$, Steffen Udluft$^1$, and Thomas Runkler$^{1,2}$
\thanks{
The project this report is based on was supported with funds from the German Federal Ministry of Education and Research under project number 01IS24087A. 
The sole responsibility for the report's contents lies with the authors.
}
%
\vspace{.3cm}\\
%
1- Siemens AG, Munich, Germany
%
\vspace{.1cm}\\
2- TU Munich (TUM), Munich, Germany
}

\maketitle

\begin{abstract} 
This paper investigates the instability of Q-learning in continuous environments, a challenge frequently encountered by practitioners.
Traditionally, this instability is attributed to bootstrapping and regression model errors. 
Using a representative reinforcement learning benchmark, we systematically examine the effects of bootstrapping and model inaccuracies by incrementally eliminating these potential error sources. 
Our findings reveal that even in relatively simple benchmarks, the fundamental task of Q-learning -- iteratively learning a Q-function from policy-specific target values -- can be inherently ill-posed and prone to failure. 
These insights cast doubt on the reliability of Q-learning as a universal solution for reinforcement learning problems.

\end{abstract}

\section{Introduction \& related work}  
\label{section:intro}

Q-learning \cite{watkins1992q} is one of the most fundamental reinforcement learning (RL) concepts making it the foundation of many popular RL algorithms.
However, from the perspective of an industrial practitioner it often falls short in terms of learning stability.
The core idea of Q-learning is to iteratively update Q-values using the Bellman equation.
While this approach works well for table-based Markov Decision Processes (MDPs), many relevant MDPs involve a continuous state space, necessitating the use of a function approximator to learn the Q-function. 
Additionally, the frequent requirement to learn offline means that Q-learning combines bootstrapping, off-policy learning, and function approximation. 
This combination, known as the \emph{deadly triad} \cite{sutton2018introduction,vanhasselt2018deepreinforcementlearningdeadly}, presents significant challenges.

In this paper, we investigate how to mitigate the known issues of Q-learning on a representative RL benchmark with a continuous state space, and why achieving stability remains challenging. 
We begin by using the well-established model-free Q-learning algorithm Neural Fitted Q Iteration (NFQ) \cite{riedmiller2005neuralnfqShort} as a baseline. 
Next, we eliminate bootstrapping by employing the model-based boot-strapping-free NFQ (BSF-NFQ) \cite{wissmann2024short}, and finally, we address model inaccuracies by utilizing the real environment dynamics. 
Throughout our study, we compare the robustness of policy learning, demonstrating significant improvements, but we are unable to completely eliminate instability.

Finally, we show that fitting the targets can result in performance variability among policies. 
By visualizing the true Q-function, we reveal a structure that cannot be accurately approximated using a neural network (NN), rendering the Q-learning task ill-posed. 
Furthermore, we demonstrate that the problem is induced by the definition of the 
MDP 
and not the algorithm itself. 
Consequently, this issue affects not only Q-learning but also other methods that rely on sample-based evaluation of Q-values.

\section{Experimental setup}  
\label{section:setup}

This paper explores
stability issues commonly encountered in continuous state MDPs like inverted pendulum, acrobot and hopper \cite{wang2023fractal}.
Due to illustration purposes, 
we perform the experiments on the iconic cart-pole balancing benchmark.

The state space is four-dimensional, \textit{i.e.}, position $x$, velocity $\dot{x}$, angle $\theta$, and angular velocity $\dot{\theta}$.
The dataset $D$ has been generated by a random policy on the gym environment \textit{CartPole-v1} from the RL benchmark library \textit{Gymnasium}\footnote{\url{https://gymnasium.farama.org}}. 
$D$ consists of 20{\small,}000 observation tuples of form $(s_t,a_t,s_{t+1},r_t)$.
Depending on state $s_t$ and action $a_t$, the system transitions to the next state $s_{t+1}$ and the agent receives a real-value reward $r_t\in\mathbb{R}$.

For the reward, we define a function that assigns 1 for an upright pole with the cart in the center and decreases quadratically along cart position and pole angle relative to their termination bounds, \textit{i.e.}, $r = (1 - (x/2.4)^2 + 1 - (\theta/0.2095)^2)/2$.

In our experiments, Q-functions were approximated using NNs in a supervised learning manner using the Adam optimizer with learning rate 0.01, mini batch size 100 and mean squared error as loss function.
The dataset was split into blocks of $70 \%$ and $30 \%$ (training and validation, respectively).
Like in \cite{wissmann2024short}, NNs with a 5-64-1 architecture were used with state-action tuples as input 
and ReLU non-linearity.
Early stopping was employed against overfitting, halting training when no improvement of the validation error for 50 epochs was made and the best parameters found so far persisted.

\section{From bootstrapping to real Q-values} 
\label{section:BSF}

The NFQ algorithm fits with an NN iteratively the targets
\begin{equation}
    Q_{i+1}(s_t,a_t) \leftarrow r_t+\gamma \max_{a_{t+1}} Q_i(s_{t+1},a_{t+1}).
    \label{eq:nfq}
\end{equation}

In \cite{wissmann2024short}, an
alternative
algorithm called BSF-NFQ was introduced. 
Here, the target calculation utilized model-based rollouts 
\begin{equation}
    {\tilde{Q}}^{\pi}_{\text{MB}}(s, a) = R(s,a,\tilde{s}_1) + \sum^{K-1}_{k=1} \gamma^{k} R(\tilde{s}_{k},\pi(\tilde{s}_{k}),\tilde{s}_{k+1}),
    \label{eq:rollout_return}
\end{equation}
where $\tilde{s}_{k+1} = M(\tilde{s}_k, \pi(\tilde{s}_{k}))$, and
is calculated for a transition model $M$ and a reward model $R$ which both can be learned from the offline dataset. 
BSF-NFQ increased robustness significantly by calculating targets with 
\begin{equation}
    Q_{i+1}(s_t,a_t) \leftarrow r_t+ \gamma {\tilde{Q}}^{\pi}_{\text{MB}}(s_{t+1}, \pi(s_{t+1})), \text{ with } \pi(s)=\argmax_a Q_i(s,a).
    \label{eq:bsf}
\end{equation}

\begin{figure}[ht!]
    \begin{subfigure}{1.\textwidth}
        \includegraphics[width=1.\textwidth]{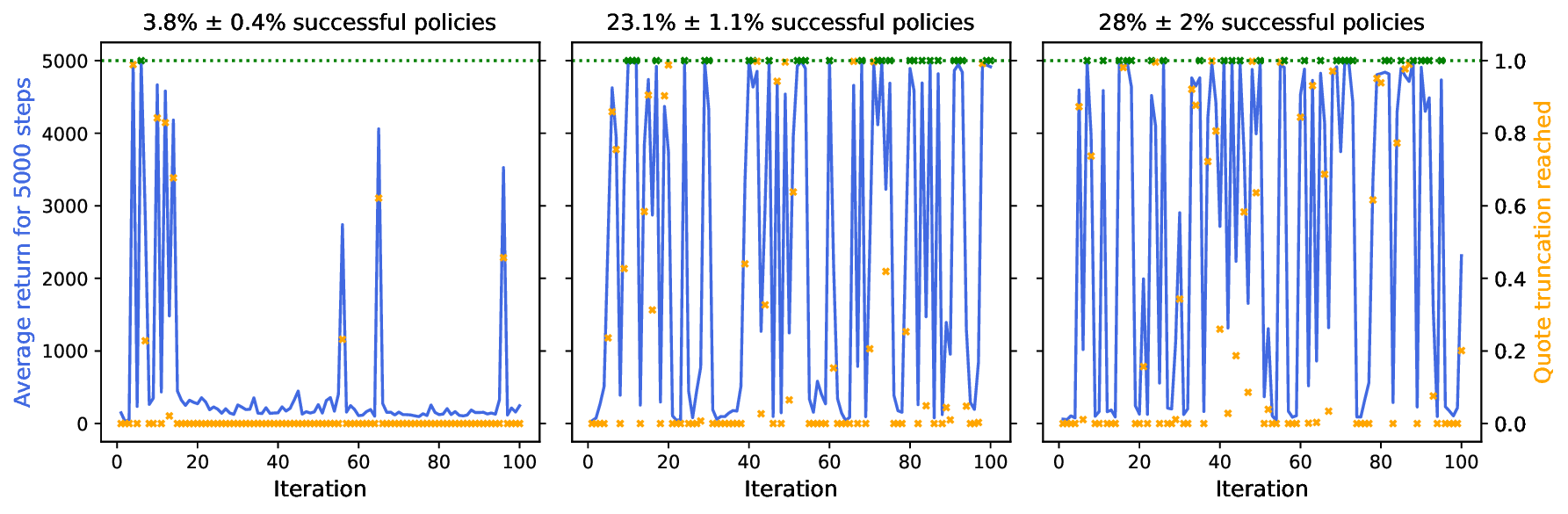}
        \subcaption*{\hspace{1.2cm}(a) NFQ \hspace{1.8cm} (b) BSF-NFQ \hspace{1.cm} (c) BSF-NFQ-real-dyn}
    \end{subfigure}
    \caption{
    Iteration-wise policy performance averaged over 1{\small,}000 gym environment episodes.
        Blue lines represent the average return over 1{\small,}000 episodes each with 5{\small,}000 steps.
        Cross markers depict the quote of episodes reaching 5{\small,}000 steps.
        Green markers represent iterations where \textit{successful} policies have been found. 
        }\label{Fig:PolicyPerf}
\end{figure}

As a result, the instability across iterations was reduced significantly but not fully resolved (see Figure~\ref{Fig:PolicyPerf}~(b)). 
After learning \textit{successful} policies, \textit{i.e.}, policies that balance successfully 
for 1{\small,}000 starting states
for at least 5{\small,}000 steps, follow-up iterations yielded bad policies again.
This gives rise to the question if this is due to learning imperfect transition models $M$ on the dataset. 

To eliminate this potential source of error, we replaced $M$ with the available benchmark transition equations and calculated the true Q-value targets using Eq.~\eqref{eq:bsf}.
A rollout horizon of $K=1,000$ and discount factor $\gamma=0.99$ was chosen, resulting in a truncation error of $\epsilon_{\text{trunc}}<0.005$.

Training 100 BSF-NFQ iterations on true Q-values yielded on average in $28\%$ 
of the iterations successful policies (see Figure~\ref{Fig:PolicyPerf}~(c)). 
Replacing the bootstrapping in NFQ by model-based policy rollout state-value
estimates (BSF-NFQ) dramatically improved the robustness of the learning algorithm and using the real dynamics (BSF-NFQ-real-dyn) improved this even further.
The reoccurring performance drops become rarer as depicted in Figure~\ref{Fig:PolicyPerf}, but still persist,
and thus, suggest a fundamental problem with Q-learning.

\section{Observing policy performance instabilities}
\label{section:fitting}

\begin{figure}[ht!]
        \includegraphics[width=1.\textwidth]{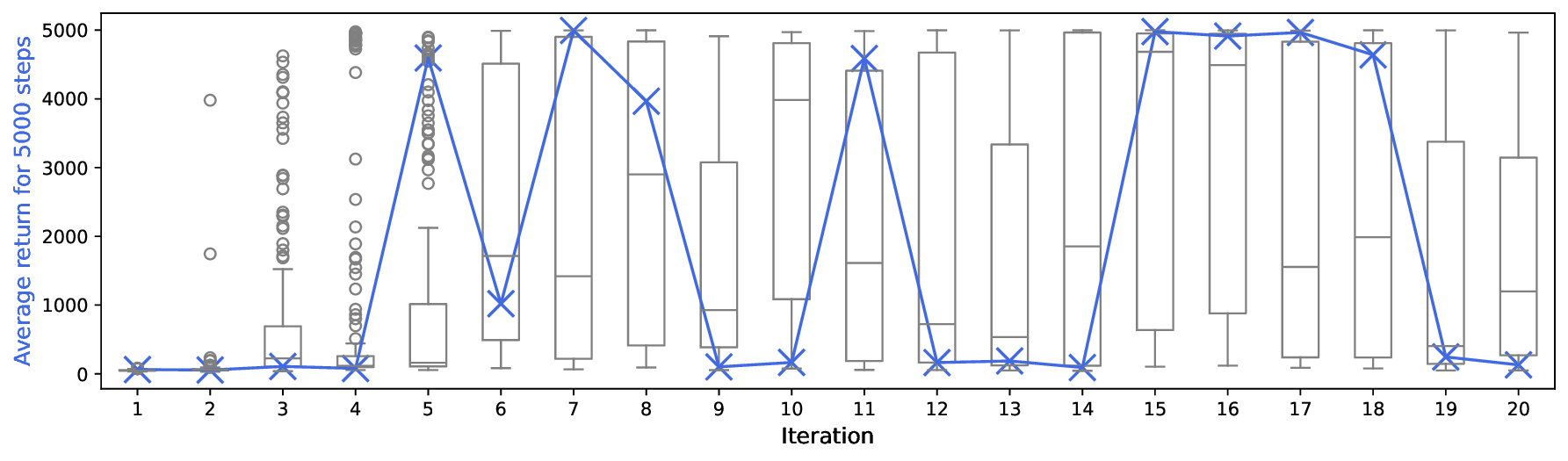}
    \caption{
        Iteration-wise policy performance averaged over 1{\small,}000
        episodes. Average return of the original iterations are depicted in blue. The boxplots visualize policy performance results 
        for retraining with saved Q-value targets on 100 seeds.
    }
    \label{Fig:DirectQ}
\end{figure}

Now that we have eliminated any potential transition model error by using the true transition and reward equations, the next step is to investigate the Q-function fitting process itself.
To gain insights into the observed instability, \textit{e.g.}, the successful policy in iteration 18 yielded a bad policy in iteration 19, we save the targets calculated with Eq.~\eqref{eq:bsf} based on the policy in iteration $i$ and retrain iteration $i+1$ with NNs initialized with different seeds.
The resulting policies were tested with respect to their performance and the corresponding box-plots for the first $20$ iterations are depicted in Figure~\ref{Fig:DirectQ}.

Surprisingly, even within one iteration, the retrained policies show a large spread in performance although they use the same Q-value targets.
Examining
potential indicators throughout the training, such as training and validation error, showed no correlation with the resulting policy performance.
Similarly, adjusting hyperparameters like total epochs to train or patience for early stopping also showed no significant impact on the performance variability.  
Furthermore, the large spread is observed for nearly all iterations.
This points to an inherent issue with the Q-value targets themselves since they do not necessarily seem to be useful to learn if we want to learn a better policy.

\section{
An ill-posed learning task}
\label{section:targets}

\begin{figure}[ht!]
    \begin{subfigure}{0.5\textwidth}
        \includegraphics[width=1\textwidth]{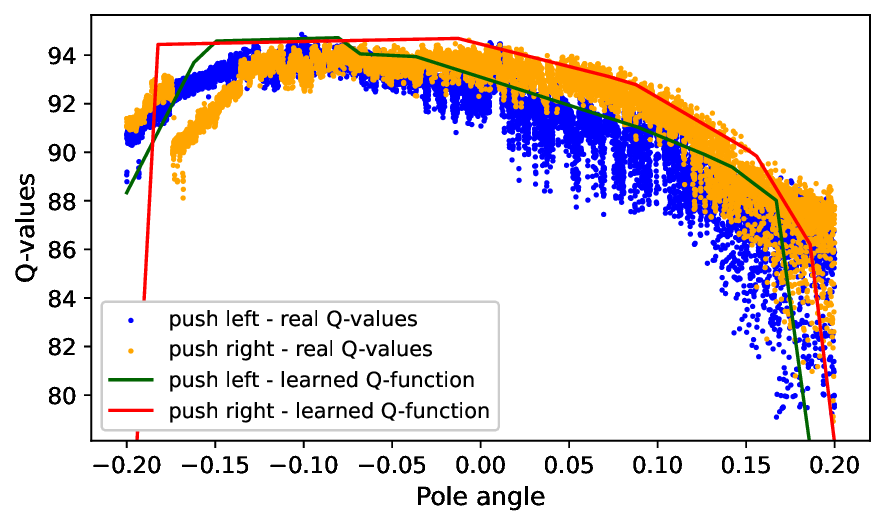}
        \subcaption{Policy from iteration 18}
        \label{fig:iter18}
    \end{subfigure}
    \begin{subfigure}{0.5\textwidth}
        \includegraphics[width=1\textwidth]{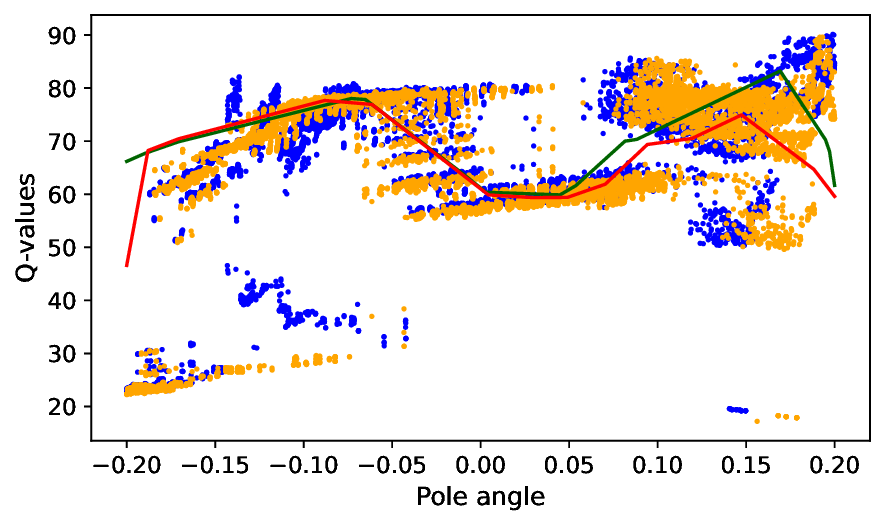}
        \subcaption{Policy from iteration 19}
        \label{fig:iter19}
    \end{subfigure}
    \begin{subfigure}{0.5\textwidth}
        \includegraphics[width=1\textwidth]{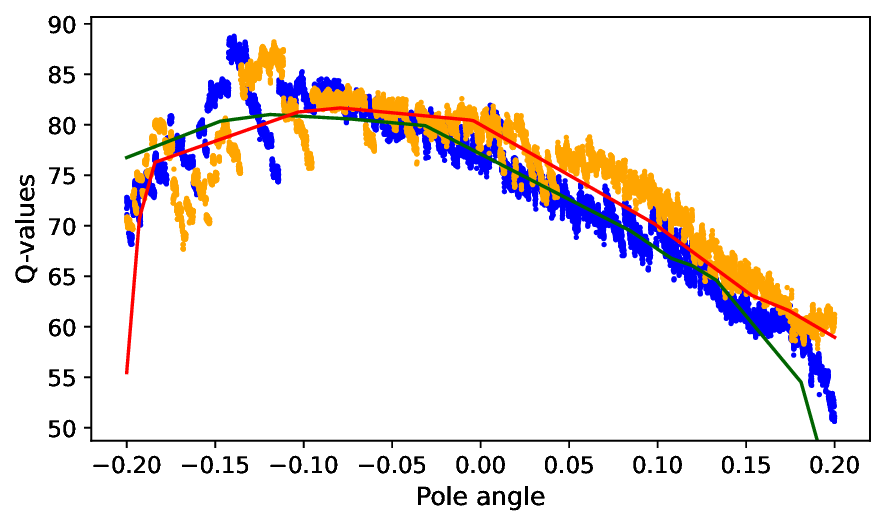}
        \subcaption{$\epsilon$-greedy policy from iteration 18}
        \label{fig:stoch_rollout}
    \end{subfigure}
    \begin{subfigure}{0.5\textwidth}
        \includegraphics[width=1\textwidth]{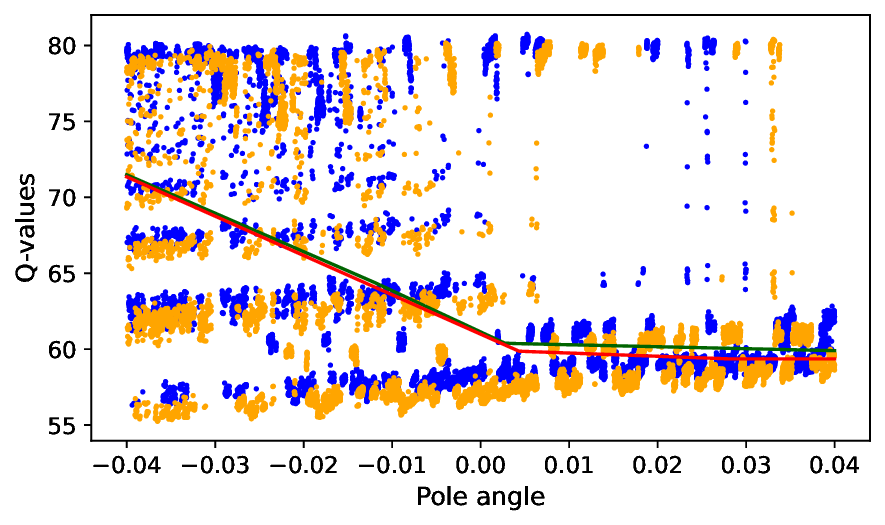}
        \subcaption{Policy from iteration 19 - magnified}
        \label{fig:zoomed}
    \end{subfigure}
    \caption{
        Q-function values along 10{\small,}000 different pole angle values 
        with cart position, cart velocity and pole velocity fixed at $0.0$. To calculate the rollouts, policies that are greedy with respect to their learned Q-value approximation were used, or for (c) $\epsilon$-greedy with an $\epsilon =0.05$. 
        Note that the plots display only a single Q-value for each corresponding angle. Therefore, the significant differences between neighboring angle values indicate function discontinuities.
    }
    \label{Fig:BSFQ}
\end{figure}

To explore why learning Q-values even for a relatively simple benchmark like cart-pole leads to vastly different Q-function approximations in terms of balancing performance, we conducted an in-depth analysis of the structure of the actual Q-values. 
Figure~\ref{Fig:BSFQ} illustrates slices through the true Q-function by holding the three state space variables $x$, $\dot{x}$, and $\dot{\theta}$ constant at 0, and plotting the Q-values across 10,000 gridded values of the pole angle $\theta$. 

Figures~\ref{Fig:BSFQ}~(a) and (b) display the Q-values from iterations 18 and 19 of the Q-learning run shown in Figure~\ref{Fig:DirectQ}, respectively. 
Notably,
the policy resulting from iteration 18 successfully balanced the pole for all starting states. 
However, in the subsequent iteration 19, the policy's performance completely collapsed.

The figures reveal the highly discontinuous structure of the true Q-values which is particularly evident in the magnified view depicted in Figure~\ref{Fig:BSFQ}~(d).
We want to emphasize that the substantial differences 
in the true Q-values
between adjacent angle values are not caused by noise, as the policy rollouts in Figures~\ref{Fig:BSFQ}~(a), (b), and (d) are entirely deterministic.

Learning a subsequent 
Q-function
on samples of these true Q-values fails to capture the structure completely and results in the NN approximating an average over the samples, which is illustrated by the line plots in Figure~\ref{Fig:BSFQ}. 
This loss of information is the primary source of error for the policy performance instability. 

The subsequent policy can encounter states where the true Q-value for the optimal action was underestimated while the true Q-value for the other action was overestimated. 
This effect can result in the subsequent policy to take a suboptimal action and
in 
case of 
the cart-pole benchmark,
taking several suboptimal actions consecutively
can result in a quickly terminating trajectory.
Thus,
the policy in iteration $i+1$ can be significantly worse than the policy from iteration $i$ that was used to calculate the targets.

The effect of calculating the expected rollout stochastically with an $\epsilon$-greedy policy, as depicted in Figure~\ref{Fig:BSFQ}~(c), shows that the randomness reduces the size of discontinuities, but does not remove them.

\begin{figure}[ht!]
    \begin{subfigure}{0.5\textwidth}
        \includegraphics[width=1\textwidth]{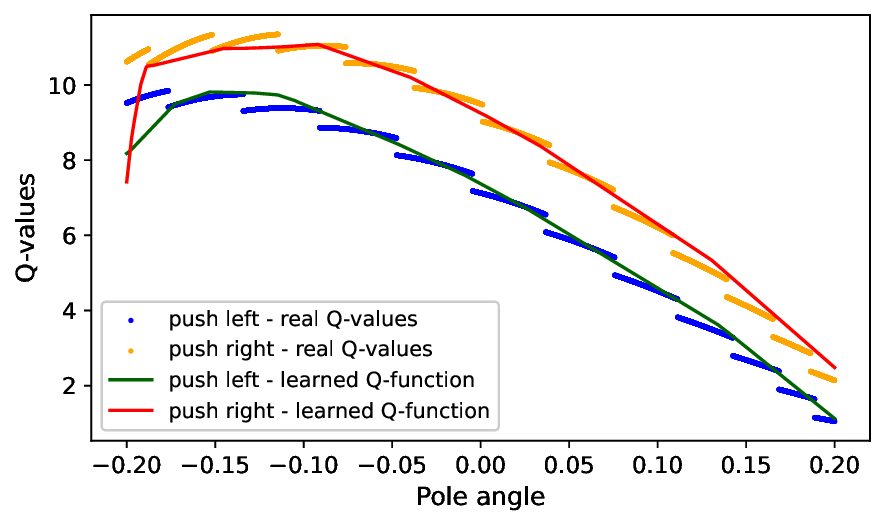}
        \subcaption{Policy that always pushes left}
        \label{fig:pushleft_q}
    \end{subfigure}
    \begin{subfigure}{0.5\textwidth}
        \includegraphics[width=1\textwidth]{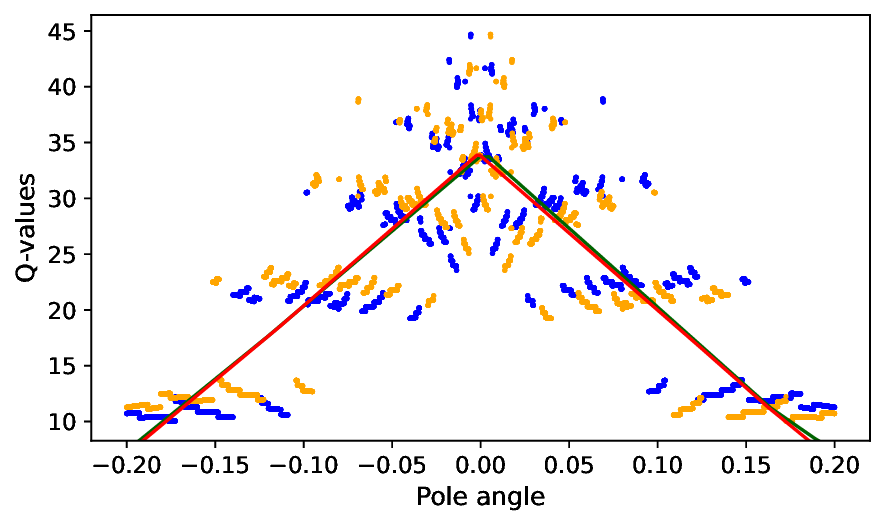}
        \subcaption{Policy against pole angle}
        \label{fig:antipole_q}
    \end{subfigure}
    \caption{
        Q-function values along 10{\small,}000 different pole angle values with cart position, cart velocity and pole velocity fixed at 0.0.
    }
    \label{Fig:SimplePolQ}
\end{figure}

To demonstrate the observed discontinuities are fundamental and not solely due to specific policies in Q-learning, Figure~\ref{Fig:SimplePolQ} shows the Q-function for two simple policies.
On the left, the Q-function for the policy that pushes left is shown, 
on the right, 
for the policy acting against the pole's angle is depicted.
In both cases, the discontinuities manifest, indicating they are a fundamental characteristic of the underlying MDP rather than an artifact of the policy design.

These 
results 
suggest that the discontinuities can already occur if the state space of the MDP is continuous. 
We argue that discontinuities in the Q-function or return values affect any method that uses them in a sample-based manner.
This includes all methods using function approximators that have been derived from Q-learning or are based on the evaluation of the return on a sample, as well as that of offline policy evaluation.
Although Q-learning can yield desired policies, discontinuities can lead to performance collapse and divergent behavior. 
Trying to fit an NN with samples from a discontinuous function makes the problem ill-posed in the first place.

\section{Conclusion}
\label{section:Conclusion}

In this paper, we identified a fundamental issue with estimating Q-values and return values. 
While estimating Q-values at isolated points can be relatively effective, using a function approximator to learn them presents significant challenges. 
We demonstrated the dramatic impact that discontinuities in the Q-function can have on Q-learning, potentially causing a collapse in the quality of the learned policies. 
Our findings illustrate that this problem can emerge even in simple MDPs with continuous state spaces.


\begin{footnotesize}

\bibliographystyle{unsrt}
\bibliography{references.bib}

\end{footnotesize}


 \end{document}